\documentclass[10pt,twocolumn,letterpaper]{article}

\usepackage[final]{cvpr}      
\usepackage{bm}
\usepackage{tabularx}
\usepackage{multirow}
\usepackage{colortbl}
\definecolor{cvprblue}{rgb}{0.21,0.49,0.74}
\usepackage[pagebackref,breaklinks,colorlinks,allcolors=cvprblue]{hyperref}

\def\paperID{10302} 
\def\confName{CVPR}
\def\confYear{2025}

\title{$\mathbf{\Phi}$-GAN: Physics-Inspired GAN for Generating SAR Images Under Limited Data}

\author{
Xidan Zhang\textsuperscript{1,\dag},  
Yihan Zhuang\textsuperscript{1,\dag},  
Qian Guo\textsuperscript{2},
Haodong Yang\textsuperscript{1},
Xuelin Qian\textsuperscript{1},
Gong Cheng\textsuperscript{1}, \\
Junwei Han\textsuperscript{1},  
Zhongling Huang\textsuperscript{1}  
\thanks{\textsuperscript{1}School of Automation, Northwestern Polytechnical University. 
\textsuperscript{2}College of Electronic and Information Engineering, Nanjing University of Aeronautics and Astronautics. \\
{\dag}: Equal contribution. \\ *: Corresponding author: huangzhongling@nwpu.edu.cn. Other contacts: zhangxidan@mail.nwpu.edu.cn; zhuangyihan@mail.nwpu.edu.cn
}}

\begin{document}
\maketitle
\begin{abstract}
Approaches for improving generative adversarial networks (GANs) training under a few samples have been explored for natural images. However, these methods have limited effectiveness for synthetic aperture radar (SAR) images, as they do not account for the unique electromagnetic scattering properties of SAR. To remedy this, we propose a physics-inspired regularization method dubbed $\Phi$-GAN, which incorporates the ideal point scattering center (PSC) model of SAR with two physical consistency losses. The PSC model approximates SAR targets using physical parameters, ensuring that $\Phi$-GAN generates SAR images consistent with real physical properties while preventing discriminator overfitting by focusing on PSC-based decision cues. To embed the PSC model into GANs for end-to-end training, we introduce a physics-inspired neural module capable of estimating the physical parameters of SAR targets efficiently. This module retains the interpretability of the physical model and can be trained with limited data. We propose two physical loss functions: one for the generator, guiding it to produce SAR images with physical parameters consistent with real ones, and one for the discriminator, enhancing its robustness by basing decisions on PSC attributes. We evaluate $\Phi$-GAN across several conditional GAN (cGAN) models, demonstrating state-of-the-art performance in data-scarce scenarios on three SAR image datasets.
\end{abstract}    
\section{Introduction}
\label{sec:intro}

Synthetic Aperture Radar (SAR), an active sensor that transmits and receives high-frequency microwaves, is vital in remote sensing due to its all-day, all-weather imaging capabilities \cite{datcu2023explainable}. The unique electromagnetic scattering properties and the labor-intensive annotation of less interpretable SAR images limit the availability of large-scale datasets, spurring interest in SAR image generation with deep learning models such as generative adversarial networks (GANs) \cite{goodfellow2020generative}. While GANs have made significant strides in generating realistic images, their inherent instability, exacerbated by limited data, often leads to mode collapse and discriminator overfitting. Techniques such as data augmentation \cite{karras2020training,zhao2020differentiable,jiang2021deceive}, regularization \cite{fang2022diggan,zhang2019consistency,zhaoImprovedConsistencyRegularization2021,tseng2021regularizing}, and latent code optimization \cite{zhengWhereMySpot2023,hong2022deltagan,xieLearningMemorizeFeature2022} have been developed to improve GAN stability and have shown success on small-scale datasets like subsets of CIFAR-10/-100 \cite{krizhevsky2009learning} and ImageNet \cite{deng2009imagenet}. However, these methods face distinct challenges when applied to the SAR domain, due to the extremely severe data scarcity problem and its special electromagnetic scattering characteristics.

In practical applications, only a few dozen SAR image per category with limited look angles are available \cite{guoCausalAdversarialAutoencoder2023}, as demonstrated in Figure \ref{fig:intro}a. Additionally, SAR targets exhibit unique electromagnetic scattering characteristics that vary significantly with the sensor’s look angle, known as the “azimuth angle” \cite{songLearningGenerateSAR2022}. Figure \ref{fig:intro}b illustrates the large discrepancy between "target rotation" and "image rotation" in a SAR image. This variability makes standard augmentation techniques, such as rotations commonly used for natural images, ineffective for SAR targets. Consequently, many traditional data augmentation methods \cite{karras2020training,zhao2020differentiable,zhaoImprovedConsistencyRegularization2021} fail in this context. Existing regularization techniques \cite{fang2022diggan,zhang2019consistency,zhaoImprovedConsistencyRegularization2021,tseng2021regularizing}, while sometimes effective, do not account for SAR’s physical scattering properties, resulting in suboptimal generation performance and limited physical consistency.

To address these challenges, we propose leveraging the ideal point scattering center (PSC) model to enhance GAN training for SAR image generation. The PSC model simplifies complex radar-object interactions by representing each SAR target as a collection of discrete point scatterers \cite{huang2024physics}. By estimating the physical parameters of SAR targets using the PSC model, we ensure that generated SAR images maintain physical consistency with real ones and guide the discriminator to make decisions based on electromagnetic features. However, parameter estimation for the PSC model is often computationally expensive \cite{OMP,LIHT,OMP_RELAX}, making efficient integration with GANs challenging. We found that this issue can be addressed by formulating the parameter estimation as a sparse reconstruction problem, which can be solved through an iterative optimization method \cite{yang2024interpretable,EMI-Net}. This insight enables the development of a neural module for efficient implementation.

Building on these findings, we propose $\Phi$-GAN, a novel SAR image generation method tailored for data-scarce scenarios. $\Phi$-GAN enhances existing cGAN training by integrating the PSC model to penalize physical inconsistencies in generated SAR images and refine the discriminator’s decision-making process to prevent overfitting. Specifically, we introduce a neural module that unrolls the Half-Quadratic Splitting (HQS) method to estimate SAR target physical parameters based on the PSC model. This module, with only a few learnable parameters, can be effectively optimized with limited data. By incorporating the frozen physics-aware neural module and the PSC model into the GAN framework, we propose two physical loss functions: one to guide the generator in producing physically consistent results, and another to constrain the discriminator to distinguish between real and synthetic images using key electromagnetic features derived from the PSC model, thus reducing overfitting.

To the best of our knowledge, $\Phi$-GAN is the first end-to-end physics-aware regularization method for SAR image generation that integrates the electromagnetic scattering model into GANs to enforce physical consistency and enhance generalization. Extensive experiments and comparisons with state-of-the-art GAN training techniques demonstrate that $\Phi$-GAN achieves faster convergence in generating crucial electromagnetic scattering features and ensures more stable refinement of generated details.

\noindent \textbf{Contributions.} The main contributions of this paper are as follows: \textbf{(1)} We demonstrate that integrating the PSC model of SAR into existing GAN frameworks significantly enhances their stability and generalization in data-scarce scenarios for SAR image generation. \textbf{(2)} We propose a novel $\Phi$-GAN, which incorporates a physics-inspired neural module for PSC parameter inversion and two specialized physical losses to effectively regularize GAN training. \textbf{(3)} We show that $\Phi$-GAN is adaptable to various conditional GAN architectures, enabling robust performance across diverse SAR image generation tasks.

\section{Related Work}
\label{sec:relwork}

\subsection{Training GANs with Limited Data}

GAN training is inherently unstable, and limited data exacerbates these challenges, making the process more unpredictable. Solutions include data augmentation \cite{karras2020training,zhao2020differentiable,jiang2021deceive,tran2021data,huang2022masked,dai2022adaptive}, regularization \cite{fang2022diggan,zhang2019consistency,zhaoImprovedConsistencyRegularization2021,tseng2021regularizing}, and latent space optimization \cite{zhengWhereMySpot2023,hong2022deltagan,xieLearningMemorizeFeature2022}.

\noindent \textbf{Data Augmentation.} Apart from some widely-applied general data augmentation methods, strategies like adaptive discriminator augmentation (ADA) \cite{karras2020training}, Differentiable Augmentation (DiffAugment) \cite{zhao2020differentiable}, and Adaptive Pseudo Augmentation \cite{jiang2021deceive} are developed for GAN training.

\noindent \textbf{Regularization.} Various studies have focused on enhancing the discriminator through regularization terms such as weight and gradient penalties \cite{fang2022diggan,tseng2021regularizing,kim2022feature}. Consistency regularization with data augmentation has been introduced to GANs to improve training stability. It penalizes the discriminator’s sensitivity to semantic-preserving perturbations introduced through data augmentation \cite{zhang2019consistency,zhaoImprovedConsistencyRegularization2021, tseng2021regularizing}.



\noindent \textbf{Latent space optimization.} Some researches have focused on optimizing the latent space to better represent unseen classes \cite{zhengWhereMySpot2023}. Other studies aim to disentangle class-relevant and class-independent representations within the latent space, enabling diverse image generation by modifying class-independent codes even with limited samples \cite{zhengWhereMySpot2023,hong2022deltagan,xieLearningMemorizeFeature2022}.

While effective for natural images, these methods often underperform in generating SAR images with limited data and may even degrade performance. Although recent studies have incorporated geometry, physics and functionality of targets in the real-world to guide generative model training \cite{mezghanniPhysicallyawareGenerativeNetwork2021,ni2024phyrecon,pizzati2023physics}, few have considered the unique characteristics of electromagnetic imaging. This motivates us to introduce a specialized GAN framework that leverages physical properties of SAR.

\subsection{SAR Image Generation}

\noindent \textbf{Controllable generation based on cGAN.} Generating SAR images with a full-range of azimuth angles from limited observations has attracted most attention in recent studies, as demonstrated in Figure \ref{fig:intro}a. Conditional GANs are the most preferred choice. The input conditions include target class \cite{shiISAGANHighFidelityFullAzimuth2022,duHighQualityMulticategorySAR2022,qinTargetSARImage2022,caoLDGANSyntheticAperture2020,caoDemandDrivenSARTarget2022,zhuLIMEBasedDataSelection2022,zouMWACGANGeneratingMultiscale2020,sunAttributeGuidedGenerativeAdversarial2023,ohPeaceGANGANBasedMultiTask2021,guoCausalAdversarialAutoencoder2023}, azimuth angle \cite{shiISAGANHighFidelityFullAzimuth2022,sunAttributeGuidedGenerativeAdversarial2023,ohPeaceGANGANBasedMultiTask2021,songLearningGenerateSAR2022,guoCausalAdversarialAutoencoder2023}, instance semantic map \cite{guoSARImageData2022}, target location \cite{juSARImageGeneration2024}, etc. Controllable SAR image generation methods based on cGANs typically encode condition factors as input and use multiple auxiliary classifiers to assess whether the generated images align with the input conditions. There are multiple strategies to input the conditions, such as concatenating them with noise vector in the input layer \cite{qinTargetSARImage2022,caoLDGANSyntheticAperture2020,caoDemandDrivenSARTarget2022,zhuLIMEBasedDataSelection2022,zouMWACGANGeneratingMultiscale2020,sunAttributeGuidedGenerativeAdversarial2023,ohPeaceGANGANBasedMultiTask2021,guoCausalAdversarialAutoencoder2023}, applying a projection module and multiplying with latent features \cite{shiISAGANHighFidelityFullAzimuth2022,duHighQualityMulticategorySAR2022}, and concatenating with multi-scale features \cite{songSARImageRepresentation2019,songLearningGenerateSAR2022}.

\noindent \textbf{Solutions designed for SAR.} Conventional data augmentation and regularization methods for natural images have been applied to SAR \cite{luo2020synthetic,juSARImageGeneration2024,zouMWACGANGeneratingMultiscale2020}, but their performance diminishes when training with only a few dozen samples per category. Several enhanced GAN architectures have been developed to reduce model complexity \cite{songLearningGenerateSAR2022,shiISAGANHighFidelityFullAzimuth2022} or to improve the feature representation of the discriminator \cite{zouMWACGANGeneratingMultiscale2020}. Some methods have focused on optimizing latent code representation to boost the diversity and quality of generated images \cite{guoCausalAdversarialAutoencoder2023,huFeatureLearningSAR2021}. Specifically, studies have identified intrinsic data issues in SAR, such as speckle noise contamination \cite{wangSyntheticApertureRadar2019, guoSyntheticApertureRadar2017} and varying scattering characteristics with target pose \cite{songLearningGenerateSAR2022}, which can lead to mode collapse and overfitting in GAN models. Consequently, specific solutions tailored to SAR characteristics have been proposed. To mitigate the negative influence of speckled clutter on discriminator’s decision-making, prior statistical distributions have been used to guide GANs toward learning SAR images with reduced speckle noise \cite{wangSyntheticApertureRadar2019,guoSyntheticApertureRadar2017} or achieving a desired gray-level distribution \cite{marmanis2017artificial}. Another significant challenge is the limited scientific validity of SAR images generated by AI models. Data-driven generative models often lack awareness of the physical principles underlying SAR imaging, leading to inconsistency with electromagnetic properties and limited generalization capacity \cite{datcu2023explainable,10752552}. Electromagnetic scattering models, with their strong theoretical foundation, are highly beneficial for guiding generative model training \cite{fuReconstruct3DGeometry2021}. However, integrating them with generative AI models for end-to-end training is often expensive and difficult. In our work, a new solution is proposed to develop a physics-inspired module inserted into existing GANs, that inherits the physical interpretability of the electromagnetic model while maintaining efficiency.

\section{Method}

\begin{figure*}[t]
  \centering
  \includegraphics[width=0.8\linewidth]{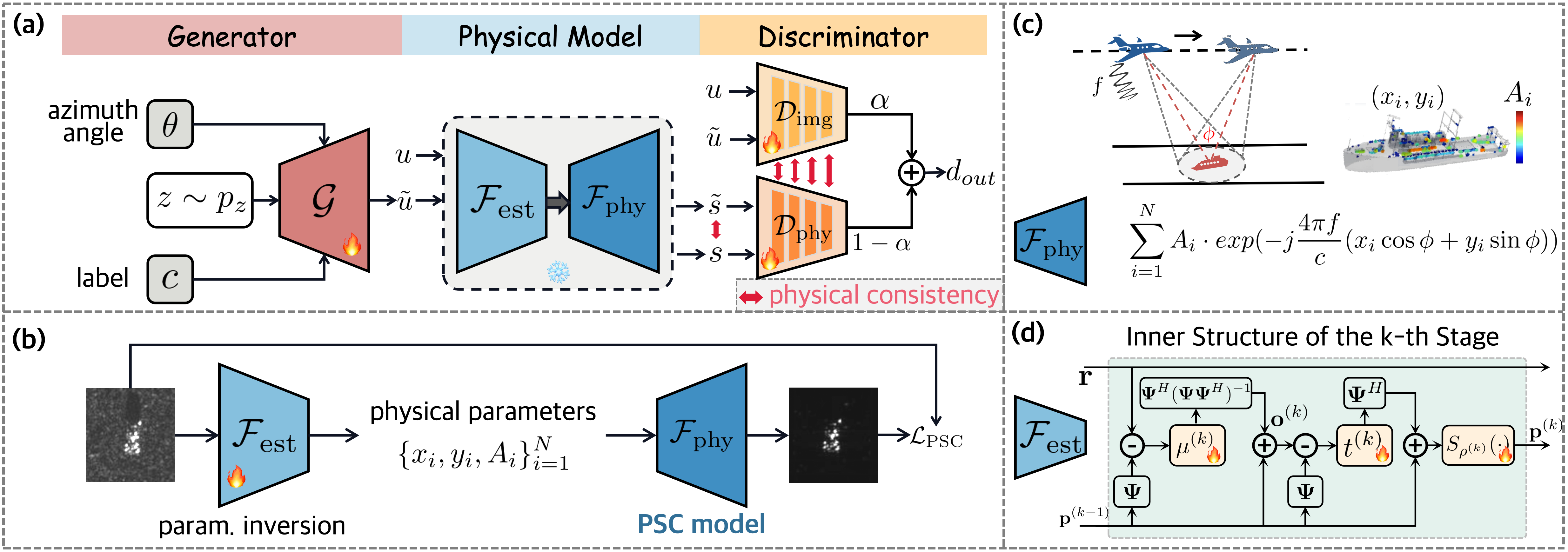}
  \caption{(a) The proposed $\Phi$-GAN framework overview. (b) The demonstration of the physics-inspired neural module ($\mathcal{F}_{\mathrm{est}}$) for physical parameter inversion and image reconstruction based on the point scattering center (PSC) model ($\mathcal{F}_{\mathrm{phy}}$). $\mathcal{F}_{\mathrm{est}}$ contains a couple of learnable parameters and $\mathcal{F}_{\mathrm{phy}}$ is an off-the-shelf reconstruction process. (c) The illustration of PSC model $\mathcal{F}_{\mathrm{phy}}$. (d) The implementation of the $k$-th stage within $\mathcal{F}_{\mathrm{est}}$.}
  \label{fig:method}
  \vspace{-6mm}
\end{figure*}

\subsection{Conditional GANs for SAR}

Conditional GANs (cGANs) are employed to generate SAR images with conditions of target label $c$ and azimuth angle $\theta$. The class information $c$ is encoded in the one-hot form. The azimuth angle $\theta$ is represented in a continuous and periodical manner with Cyclic High-frequency information-based Embedding (CHE) method \cite{guoCausalAdversarialAutoencoder2023}, denoted as:
\begin{equation}
\begin{split}
    r(\theta)= & [sin(\theta),cos(\theta),sin(2\theta),...,sin(5\theta),cos(5\theta)]
\end{split}
\end{equation}
The encoded conditions are then concatenated  with the latent code $z$ as the input of the generator, denoted as $\{z,c,r(\theta)\}$. The training objective of the cGANs can be written below:
\begin{equation}
\begin{aligned}
\label{equ:origanloss}
\mathcal{L}^D_{\mathrm{ori}} &= -{\mathbb{E}}_{u,c,\theta}[f(D(u|c,\theta))]+ {\mathbb{E}}_{z,c,\theta}[f(D(G(z,c,\theta)))] \\
\mathcal{L}^G_{\mathrm{ori}} &= -{\mathbb{\mathbb{E}}}_{z,c,\theta}[f(D(G(z,c,\theta)))]
\end{aligned}
\end{equation}
where $f(\cdot)$ is the loss function of the discriminator.

\subsection{$\Phi$-GAN Overview}

The overview of $\Phi$-GAN is illustrated in Figure \ref{fig:method}a. Unlike existing cGANs, $\Phi$-GAN integrates a physical model component, represented as $\mathcal{F}_{\mathrm{est}} \circ \mathcal{F}_{\mathrm{phy}}$, and incorporates a dual-discriminator structure comprising $\mathcal{D}_{\mathrm{img}}$ and $\mathcal{D}_{\mathrm{phy}}$. These discriminators are responsible for assessing the generated image and the output of the physical model, respectively. Specifically, $\mathcal{F}_{\mathrm{est}}$ estimates the physical parameters of an input SAR image, while $\mathcal{F}_{\mathrm{phy}}$ reconstructs its electromagnetic (EM) scattering features. The EM reconstruction outputs for real and generated SAR images are denoted as $s$ and $\tilde{s}$, respectively. $\mathcal{D}_{\mathrm{img}}$ and $\mathcal{D}_{\mathrm{phy}}$ independently discriminate between real and generated images, as well as their reconstructed EM features based on the estimated physical parameters. The final discriminator decision combines the outputs of $\mathcal{D}_{\mathrm{img}}$ and $\mathcal{D}_{\mathrm{phy}}$. The process is defined as follows:
\begin{equation}
\begin{aligned}
\label{equ:generator}
    \tilde{u} &= \mathcal{G}(z,c,r(\theta)), \\
    s,\tilde{s} &= \mathcal{F}_{\mathrm{est}} \circ \mathcal{F}_{\mathrm{phy}} (u,\tilde{u}), \\
    d_{out}(u,s) &= \alpha \mathcal{D}_{\mathrm{img}}(u) + (1-\alpha) \mathcal{D}_{\mathrm{phy}}(s)
\end{aligned}
\end{equation}
To efficiently integrate the physical model into GAN training in an end-to-end manner, we propose a physics-inspired neural module to implement $\mathcal{F}_{\mathrm{est}}$ with a few learnable parameters, as detailed in Section \ref{sec:pinn}. Additionally, we introduce two physical loss functions, $\mathcal{L}_{\mathrm{phy}}^G$ and $\mathcal{L}_{\mathrm{phy}}^D$. The purpose of $\mathcal{L}_{\mathrm{phy}}^G$ is to penalize unrealistic generated SAR images with inconsistent physical parameters, while $\mathcal{L}_{\mathrm{phy}}^D$ regularizes the discriminator to make more robust decisions based on the discriminative EM features of targets. These losses complement the original optimization objectives, as outlined in Section \ref{sec:opt}:
\begin{equation}
    \begin{aligned}
    \label{equ:oriloss}
    \mathcal{L}^G &= \mathcal{L}^G_{\mathrm{ori}} + \mathcal{L}_{\mathrm{phy}}^G, \\
    \mathcal{L}^D &= \mathcal{L}^D_{\mathrm{ori}} + \mathcal{L}_{\mathrm{phy}}^D.
    \end{aligned}
\end{equation}
\subsection{Physics-Inspired Neural Module}
\label{sec:pinn}

Based on geometric diffraction theory, the electromagnetic scattering of a target can be approximately represented as the sum of responses from multiple independent ideal PSCs at high frequencies. The response of each PSC is a function of the radar echo frequency $f$ and aspect angle $\phi$, as illustrated in Figure \ref{fig:method}c. The SAR target with $N$ PSCs can be modeled as:
\begin{equation}
\label{equ:PSC}
    E(f,\phi) = \sum\limits_{i=1}^{N} A_i \cdot \operatorname {exp}(-j\frac{4\pi f}{c}(x_i \cos{\phi} + y_i \sin{\phi})),
\end{equation}
\noindent where $E$ denotes the measurement in the frequency domain. $A_i$ and $\exp(\cdot)$ are the response intensity and the phase information of radar echos of the $i$th PSC, respectively. $x_i$ and $y_i$ decide the position of each PSC, as demonstrated in Figure \ref{fig:method}c. In this work, the physical parameters ${A_i, x_i, y_i}$ modeled with PSC are first estimated in $\mathcal{F}_{\mathrm{est}}$ and then reconstructed into images in $\mathcal{F}_{\mathrm{phy}}$. An overview of this process is provided in Figure \ref{fig:method}b.

Previous study \cite{yang2020efficient} has shown that radar echo signals exhibit sparsity in image domain, enabling the use of sparse signal representation for analysis. Thus, the PSC parameters can be estimated by optimizing the following problem:
\begin{equation}h
\label{equ:lasso}
    \mathbf{\hat{o}} = \mathop{\arg\min}\limits_{\mathbf{o}}||\mathbf{\Psi} \mathbf{o}-\mathbf{r}||_2+\lambda||\mathbf{o}||_1.
\end{equation}
The symbol $\mathbf{\Psi}$ represents a dictionary that contains the positional information of PSCs, corresponding to the exponential term in Equation (\ref{equ:PSC}). More details about $\mathbf{\Psi}$ will be provided in the supplementary materials. $\mathbf{r}$ denotes the input image after vectorization. $\mathbf{o}$ denotes a sparse vector that stores the response values at each position in the image, corresponding to $\{A_i\}_{i=1}^{N}$ in Equation (\ref{equ:PSC}). $\mathbf{\hat{o}}$ is the estimation of $\mathbf{o}$ and $\lambda$ is a hyperparameter that balances the tradeoff between data fidelity and sparsity. However, existing methods, such as HQS \cite{hqs}, often rely on iterative optimization strategies and manual hyperparameter tuning, which result in low processing speed and suboptimal accuracy. To overcome this challenge, we introduce a neural module $\mathcal{F}_{\mathrm{est}}$ that unrolls the HQS algorithm for efficient, real-time estimation of PSC parameters. This approach provides immediate feedback from the physical model, enabling end-to-end training of $\Phi$-GAN.




HQS decomposes the problem into two alternating optimization sub-problems by introducing auxiliary variables. It then gradually converges to the optimal solution of the original problem by alternately optimizing the primary and auxiliary variables. First, for the optimization problem in Equation (\ref{equ:lasso}), an auxiliary variable $\mathbf{p}$ is introduced, and the problem is reformulated as:
\begin{equation}
\label{equ:hqs}
\min_{\mathbf{o}, \mathbf{p}} \quad  \|\mathbf{r} - \mathbf{\Psi} \mathbf{o}\|_2^2 + \lambda \|\mathbf{p}\|_1, \quad \text{subject to} \quad \mathbf{o} = \mathbf{p}.
\end{equation}
\noindent Next, $\mathbf{o}$ and $\mathbf{p}$ are alternately optimized using the following optimization strategy:
\begin{equation}
\begin{split}
\mathbf{o}^{(k)} &= \arg\min_{\mathbf{o}} \left[ \frac{1}{2} \|\mathbf{r} - \mathbf{\Psi} \mathbf{o}\|_2^2 + \frac{\mu}{2} \|\mathbf{o} - \mathbf{p}^{(k-1)}\|_2^2 \right], \\
\mathbf{p}^{(k)} &= \arg\min_{\mathbf{p}} \left[ \lambda \|\mathbf{p}\|_1 + \frac{\mu}{2} \|\mathbf{o}^{(k)} - \mathbf{p}\|_2^2 \right],
\end{split}
\end{equation}
\noindent where $\mathbf{o}^{(k)}$ and $\mathbf{p}^{(k)}$ denote the values of $\mathbf{o}$ and $\mathbf{p}$ at the $k$-th iteration, respectively, and $k$ represents the iteration count. Once $\mathbf{o}^{(k)}$ is obtained, the proximal gradient algorithm can be directly applied to solve $\mathbf{p}^{(k)}$. Thus, the closed-form solutions for $\mathbf{o}^{(k)}$ and $\mathbf{p}^{(k)}$ are given as follows:
\begin{equation}
\begin{aligned}
\mathbf{o}^{(k)} &= \mathbf{p}^{(k-1)} + \mu \mathbf{\Psi}^H (\mathbf{\Psi} \mathbf{\Psi}^H)^{-1}(\mathbf{r} - \mathbf{\Psi} \mathbf{p}^{(k-1)}), \\
\mathbf{p}^{(k)} &= S_{\rho}\left(\mathbf{p}^{(k-1)} + t \mathbf{\Psi}^H (\mathbf{\Psi} \mathbf{p}^{(k-1)} - \mathbf{o}^{(k)}) \right).
\end{aligned}
\end{equation}
\noindent $S_\rho$ is the soft-thresholding function, which is expressed as $S_\rho(x) = sign(x)max(|x|-\rho,0)$. In traditional HQS, the parameters $t$, $\rho$, and $\mu$ are typically chosen based on prior information and then held fixed. In this work, we unfold HQS as a neural network and set the parameters at each stage to be learnable. The internal structure of the network stage $k$ is shown in Figure \ref{fig:method}d and can be described as follows:
\begin{equation}
\label{eq:unroll_hqs}
\begin{aligned}
\mathbf{o}^{(k)} &= \mathbf{p}^{(k-1)} + \mu^{(k)} \mathbf{\Psi}^H (\mathbf{\Psi} \mathbf{\Psi}^H)^{-1}(\mathbf{r} - \mathbf{\Psi} \mathbf{p}^{(k-1)}), \\
\mathbf{p}^{(k)} &= S_{\rho^{(k)}}\left(\mathbf{p}^{(k-1)} + t^{(k)} \mathbf{\Psi}^H (\mathbf{\Psi} \mathbf{p}^{(k-1)} - \mathbf{o}^{(k)}) \right).
\end{aligned}
\end{equation}
\noindent Building on Equation (\ref{eq:unroll_hqs}), we designed a two-stage process to construct $\mathcal{F}_{\mathrm{est}}$, resulting in a total of 6 trainable parameters. The initial values of $t^{(k)}$, $\rho^{(k)}$, and $\mu^{(k)}$ for each stage are set to $0.001$, $0.005$ and $0.001$, respectively.

For each sample, the loss is defined as follows:
\begin{equation}
\label{equ:loss}
\mathcal{L}_{\mathrm{PSC}} = ||\mathbf{r}-\mathbf{\Psi} \mathbf{o}^{(K)}||^2_2+\lambda_1 ||\mathbf{o}^{(K)} - \mathbf{p}^{(K)}||^2_2 + \lambda_2||\mathbf{p}^{(K)}||_1.
\end{equation}
$||\mathbf{r}-\mathbf{\Psi} \mathbf{o}^{(K)}||^2_2$ indicates the residual loss between the input image and the reconstructed image, where a lower value indicates a greater consistence. $\mathbf{\Psi} \mathbf{o}^{(K)}$ denotes the reconstruction procedure, i.e.  $\mathcal{F}_{\mathrm{phy}}$. $\lambda_1 ||\mathbf{o}^{(K)} - \mathbf{p}^{(K)}||^2_2$ ensures consistency between the original and auxiliary variables, while $\lambda_2||\mathbf{p}^{(K)}||_1$ enforces sparsity of the variable, demonstrating that the SAR target can be modeled with a minimal number of PSCs.
The hyperparameters $\lambda_1$ and $\lambda_2$ are empirically set to 100 and 200, respectively.

\subsection{Optimization of Generator and Discriminator}
\label{sec:opt}
\begin{table*}[t!]
\centering

\caption{Comparing SAR image generation results with varying data percentages on MSTAR dataset. ACGAN is applied as the baseline.}
\label{tab:acganmstar}
\resizebox{0.9\linewidth}{!}{
\begin{tabular}{lcccccc|cccccc}
\toprule
                                  & \multicolumn{6}{c|}{\textbf{10\% MSTAR}}                                                                                                                                                                                                              & \multicolumn{6}{c}{\textbf{5\% MSTAR}}                                                                  \\
\cmidrule{2-13}
\multicolumn{1}{c}{\multirow{-2}{*}{\textbf{Method}}} & \textbf{SSIM($\uparrow$)}                       & \textbf{VIF($\uparrow$)}                        & \textbf{FSIM($\uparrow$)}                       & \textbf{GMSD($\downarrow$)}                       & \textbf{FID($\downarrow$)}                        & \textbf{KID($\downarrow$)}                      & \textbf{SSIM($\uparrow$)}                       & \textbf{VIF($\uparrow$)}                        & \textbf{FSIM($\uparrow$)}                       & \textbf{GMSD($\downarrow$)}                       & \textbf{FID($\downarrow$)}                        & \textbf{KID($\downarrow$)}                      \\
\midrule  
\textbf{ACGAN}                    & 0.3224                                 & 0.0386                                 & 0.7432                                 & 0.1510                                 & 290.0484                                & 0.4548                                 & 0.3007                                 & 0.0315                                 & 0.7291                                 & 0.1583                                 & 340.2440                                 & 0.4713                                 \\
\textbf{+ADA\cite{karras2020training}}                     & 0.2606                                 & 0.0243                                 & 0.7171                                 & 0.1643                                 & 320.5937                                & 0.4240                                 & 0.2786                                 & 0.0235                                 & 0.7095                                 & 0.1667                                 & 364.5610                                 & 0.5304                                 \\
\textbf{+DA\cite{zhao2020differentiable}}                      & 0.2168                                 & 0.0188                                 & 0.6570                                 & 0.2018                                 & 1089.4734                               & 1.8056                                 & 0.2025                                 & 0.0188                                 & 0.6620                                 & 0.1980                                 & 976.1112                                 & 1.6584                                 \\
\textbf{+RLC\cite{tseng2021regularizing}}                     & 0.3137                                 & 0.0349                                 & 0.7362                                 & 0.1527                                 & 286.7148                                & 0.4315                                 & 0.2905                                 & 0.0293                                 & 0.7303                                 & 0.1573                                 & 375.5049                                 & 0.5698                                 \\
\textbf{+DIG\cite{fang2022diggan}}                     & 0.3279                                 & 0.0311                                 & 0.7283                                 & 0.1570                                 & 373.0722                                & 0.6243                                 & {\color[HTML]{D83931} \textbf{0.3279}} & 0.0267                                 & 0.7166                                 & 0.1598                                 & 438.7845                                 & 0.7002                                 \\
\midrule
\rowcolor{gray!20}
\textbf{+Ours}                    & {\color[HTML]{D83931} \textbf{0.3583}} & {\color[HTML]{D83931} \textbf{0.0781}} & {\color[HTML]{D83931} \textbf{0.7622}} & {\color[HTML]{D83931} \textbf{0.1385}} & {\color[HTML]{D83931} \textbf{87.2719}} & {\color[HTML]{D83931} \textbf{0.0414}} & 0.3141                                 & {\color[HTML]{D83931} \textbf{0.0547}} & {\color[HTML]{D83931} \textbf{0.7441}} & {\color[HTML]{D83931} \textbf{0.1485}} & {\color[HTML]{D83931} \textbf{130.2311}} & {\color[HTML]{D83931} \textbf{0.0861}}\\
\bottomrule 
\end{tabular}}
\end{table*}

\begin{table*}[t]
  
\noindent
\begin{minipage}[t]{0.48\linewidth}
    \centering
    \caption{Comparisons based on StyleGAN.}
    \label{tab:stylegan}
    \resizebox{\linewidth}{!}{
    \begin{tabular}{ccc|cc|cc}
    \toprule
                                      & \multicolumn{2}{c|}{\textbf{10\% MSTAR}}                                           & \multicolumn{2}{c|}{\textbf{5\% MSTAR}}                                            & \multicolumn{2}{c}{\textbf{1\% OpenSARShip}}                                     \\
    \multirow{-2}{*}{\textbf{Method}} & \textbf{FID($\downarrow$)}                          & \textbf{KID($\downarrow$)}                        & \textbf{FID($\downarrow$)}                          & \textbf{KID($\downarrow$)}                        & \textbf{FID($\downarrow$)}                         & \textbf{KID($\downarrow$)}                        \\
    \midrule   
    \textbf{StyleGAN}                 & 290.6406                                 & 0.3641                                 & 339.2788                                 & 0.4286                                 & 45.0999                                 & 0.0326                                 \\
    \midrule
    \rowcolor{gray!20}
    \textbf{+Ours}            & {\color[HTML]{D83931} \textbf{174.8329}} & {\color[HTML]{D83931} \textbf{0.1752}} & {\color[HTML]{D83931} \textbf{305.0540}} & {\color[HTML]{D83931} \textbf{0.3894}} & {\color[HTML]{D83931} \textbf{44.7829}} & {\color[HTML]{D83931} \textbf{0.0293}}\\
    \bottomrule 
    \end{tabular}}
\end{minipage}%
\hfill
\noindent
\begin{minipage}[t]{0.48\linewidth}

    \centering
    \caption{Comparisons with SAR image generation methods.}
    \label{tab:sargen}
    \resizebox{\linewidth}{!}{
    \begin{tabular}{ccc|cc|cc}
    \toprule
                                      & \multicolumn{2}{c|}{\textbf{1\% OpenSARShip}}                                     & \multicolumn{2}{c|}{\textbf{14\% SAR-Airplane}}                                   & \multicolumn{2}{c}{\textbf{10\% MSTAR}}                                         \\
    \multirow{-2}{*}{\textbf{Method}} & \textbf{FID($\downarrow$)}                         & \textbf{KID($\downarrow$)}                        & \textbf{FID($\downarrow$)}                         & \textbf{KID($\downarrow$)}                        & \textbf{FID($\downarrow$)}                         & \textbf{KID($\downarrow$)}                        \\
    \midrule 
    \textbf{CAE\cite{guoCausalAdversarialAutoencoder2023}}                      & 41.2370                                 & 0.0269                                 & 34.4476                                 & 0.0180                                 & 134.8619                                & 0.1526                                 \\
    \textbf{CVAE-GAN\cite{huFeatureLearningSAR2021}}                 & 23.9480                                 & 0.0112                                 & 34.7448                                 & 0.0137                                 & 495.9682                                & 0.5147                                 \\
    \midrule    
    \rowcolor{gray!20} \textbf{CVAE-GAN+Ours}            & 15.0530                                 & {\color[HTML]{D83931} \textbf{0.0039}} & {\color[HTML]{D83931} \textbf{17.0494}} & {\color[HTML]{D83931} \textbf{0.0021}} & 104.8976                                & 0.0662                                 \\
    \rowcolor{gray!20} \textbf{ACGAN+Ours}               & {\color[HTML]{D83931} \textbf{14.7987}} & 0.0046                                 & 19.1492                                 & 0.0046                                 & {\color[HTML]{D83931} \textbf{87.2719}} & {\color[HTML]{D83931} \textbf{0.0414}}\\
    \bottomrule 
    \end{tabular}}

\end{minipage}%

\end{table*}

SAR images differ from natural images due to speckled clutter and discrete target features. Their unique electromagnetic scattering properties pose challenges for training GANs, particularly with limited data. While current regularization methods focus on weights, gradients, and general strategies to stabilize training, we propose two physics-based loss functions for the generator and discriminator to effectively integrate SAR domain knowledge.

We propose a dual-discriminator structure as illustrated in Figure \ref{fig:method}a, and the final output of the discriminators is previously defined as $d_{out}$ in Equation (\ref{equ:generator}). With this structure, the original optimization objectives in Equation (\ref{equ:origanloss}) for the generator and discriminator are redefined as follows:
\begin{equation}
\begin{aligned}
\label{equ:newganloss}
    \mathcal{L}_{\mathrm{ori}}^G &= -{\mathbb{E}}_{z,c,\theta}[f({d}_{out}(\tilde{u},s))], \\
    \mathcal{L}_{\mathrm{ori}}^D &= -{\mathbb{E}}_{z,c,\theta}[f(d_{out}(u,s))]+{\mathbb{E}}_{z,c,\theta}[f(d_{out}(\tilde{u},\tilde{s}))].
\end{aligned}
\end{equation}

In data-scarce scenarios, the generator faces challenges in capturing the crucial electromagnetic scattering features of SAR targets, limiting its ability to produce realistic SAR images that adhere to physical principles. To address this, we propose a physical loss, $\mathcal{L}_{\mathrm{phy}}^G$, to ensure the physical consistency of the generated data. This loss imposes constraints on both the physical parameters and the latent embedding space. Specifically, it enforces that generated samples have physical parameters similar to those of real samples with the same target label and azimuth angle. Additionally, it constrains the latent feature representations of the generated image, denoted as $\{F_{\mathrm{img}}^i(\tilde{u})\}_{i=1}^M$, aligned with the real electromagnetic scattering features in the latent space, denoted as $\{F_{\mathrm{phy}}^i(s)\}_{i=1}^M$. The overall physical loss for the generator is defined as: 
\begin{equation}
    \mathcal{L}_{\mathrm{phy}}^G = \beta \mathcal{L}_{\mathrm{phy/s}}^G + \gamma \mathcal{L}_{\mathrm{phy/f}}^G, \\
\end{equation}
where $\mathcal{L}_{\mathrm{phy/s}}^G$ and $\mathcal{L}_{\mathrm{phy/f}}^G$ are image-level and feature-level constraints, respectively, and $\beta$ and $\gamma$ are the hyperparameters to control the balances of them:
\begin{equation}
\begin{aligned}
&\mathcal{L}_{\mathrm{phy/s}}^G = \mathrm{MSE}(s,\tilde{s}), \\
&\mathcal{L}_{\mathrm{phy/f}}^G = \sum_{i=1}^{M}\frac{1}{C^iH^iW^i}||F_{\mathrm{phy}}^i(s)-F_{\mathrm{img}}^i(\tilde{u})||_2.
\end{aligned}
\end{equation}
where $C^i$,$H^i$,$W^i$ represent the number of channel, height, and width of the feature map at the $i$-th stage/layer.

The discriminator often suffers from severe overfitting, leading to biased decisions based on patterns in speckle noise. To mitigate this, the proposed dual-discriminator structure introduces an auxiliary discriminator, $\mathcal{D}_{\mathrm{phy}}$, which regularizes the main image discriminator,  $\mathcal{D}_{\mathrm{img}}$, and prevents overfitting. We design a physical loss to distill the latent features of $\mathcal{D}_{\mathrm{img}}$ using those of  $\mathcal{D}_{\mathrm{phy}}$. This ensures that the image discriminator relies on key electromagnetic scattering features rather than noise patterns. The physical loss for the discriminator is defined as:
\begin{equation}
\begin{aligned}
\label{equ:phyd}
     \mathcal{L}_{\mathrm{phy}}^D = \gamma \sum_{i=1}^{M}\frac{1}{C^iH^iW^i} ( & ||F_{\mathrm{img}}^i(\tilde{u})-F_{\mathrm{phy}}^i(\tilde{s})||_2 \\ + & ||F_{\mathrm{img}}^i(u)-F_{\mathrm{phy}}^i(s)||_2 ) 
\end{aligned}
\end{equation}
By combining Equations (\ref{equ:newganloss}–\ref{equ:phyd}), the generator and discriminator are optimized in an alternating fashion.






\section{Experiments}

\subsection{Datasets and Settings}




\noindent \textbf{Datasets.} We apply three SAR image datasets: the Moving and Stationary Target Acquisition and Recognition (MSTAR) (10-class) \cite{mstar}, SAR-Airplane (2-class) \cite{Wang2022} and OpenSARShip (2-class) \cite{huang2017opensarship}. To create a data-scarce training scenario, we uniformly sample subsets from the original datasets based on the azimuth angles of the targets. Eventually, we obtain 10\% and 5\% MSTAR dataset with 237 and 121 samples for 10-class, 1\% OpenSARShip dataset with 46 samples for 2-class, and 14\% SAR-Airplane dataset with 20 samples for 2-class. More details can be found in supplementary material.


\noindent \textbf{Baseline.} We use ACGAN \cite{odena2017conditional} and StyleGAN \cite{karras2019style} as baseline models in our experiments. To tailor them to our specific task, we implement the following modifications compared to their original configurations.
\begin{itemize}
    \item[•] ACGAN \cite{odena2017conditional}: We extend the original auxiliary classifier in ACGAN by adding an angle estimator, enabling simultaneous evaluation of the input conditions for class and azimuth angle. Spectral normalization\cite{miyato2018spectral} was utilized to stable GAN training.
    \item[•] StyleGAN \cite{karras2019style}: For the generator, the embedded azimuth angle and class information are concatenated with the latent noise as input to the mapping network. Meanwhile, the conditional input of the projection discriminator \cite{miyato2018cgans} is set to the concatenation of class and azimuth angle embedding.

\end{itemize}

\noindent \textbf{Implementation details.} The hyperparameters $\alpha,\beta,\gamma$ in our proposed $\Phi$-GAN are set to 0.6, 1, 10, respectively. The learning rate is set to 0.0002 and Adam is used for optimizing GANs. At the early training stage, the generator undergoes five training iterations for each iteration of the discriminator. When training $\mathcal{F}_{\mathrm{est}}$, we adopt the AdamW optimizer with a weight delay of 0.005 and a learning rate of 0.002.

\noindent \textbf{Evaluation metrics.}
Fréchet Inception Distance (FID) \cite{heusel2017gans} and Kernel Inception Distance (KID) \cite{binkowski2018demystifying} are applied for evaluation. Additionally, for the MSTAR dataset, we incorporated four full-reference Image Quality Assessment (IQA) metrics—SSIM \cite{wang2004image}, VIF \cite{sheikh2006image}, FSIM \cite{5705575}, and GMSD \cite{6678238}—to assess the quality of the generated samples.

\subsection{Comparison with State-of-the-art Methods}

\begin{figure*}[t]
  
\begin{minipage}[t!]{0.48\linewidth}
    \vspace{-3cm}
    \centering
    \captionof{table}{Results on OpenSARShip and SAR-Airplane.}
    \label{tab:acganships}
    \resizebox{0.85\linewidth}{!}{
    \begin{tabular}{lcc|cc}
    \toprule
    \multicolumn{1}{c}{}                                  & \multicolumn{2}{c|}{\textbf{1\% OpenSARShip}}                                       & \multicolumn{2}{c}{\textbf{14\% SAR-Airplane}}                                      \\
    \cmidrule{2-5}
    \multicolumn{1}{c}{\multirow{-2}{*}{\textbf{Method}}} & \textbf{FID($\downarrow$)}                        & \textbf{KID($\downarrow$)}                   & \textbf{FID($\downarrow$)}                        & \textbf{KID($\downarrow$)}                  \\
    \midrule      
    \textbf{ACGAN}                    & 18.2855                                 & 0.0081                                 & 25.2219                                 & 0.0090                                 \\
    \textbf{+ADA\cite{karras2020training}}                    & 25.0931                                 & 0.0136                                 & 36.7829                                 & 0.0198                                 \\
    \textbf{+DA\cite{zhao2020differentiable}}                      & 37.0723                                 & 0.0203                                 & 47.8409                                 & 0.0253                                 \\
    \textbf{+RLC\cite{tseng2021regularizing}}                     & 20.5735                                 & 0.0106                                 & 31.4813                                 & 0.0157                                 \\
    \textbf{+DIG\cite{fang2022diggan}}                     & 23.5985                                 & 0.0135                                 & 21.8573                                 & 0.0066                                 \\
    \midrule
    \rowcolor{gray!20}
    \textbf{+Ours}                    & {\color[HTML]{D83931} \textbf{14.7987}} & {\color[HTML]{D83931} \textbf{0.0046}} & {\color[HTML]{D83931} \textbf{19.1492}} & {\color[HTML]{D83931} \textbf{0.0046}}\\
    \bottomrule 
    \end{tabular}}

\end{minipage}%
\hfill
\begin{minipage}[t]{0.48\linewidth}

  \centering
  \includegraphics[width=1.0\linewidth]{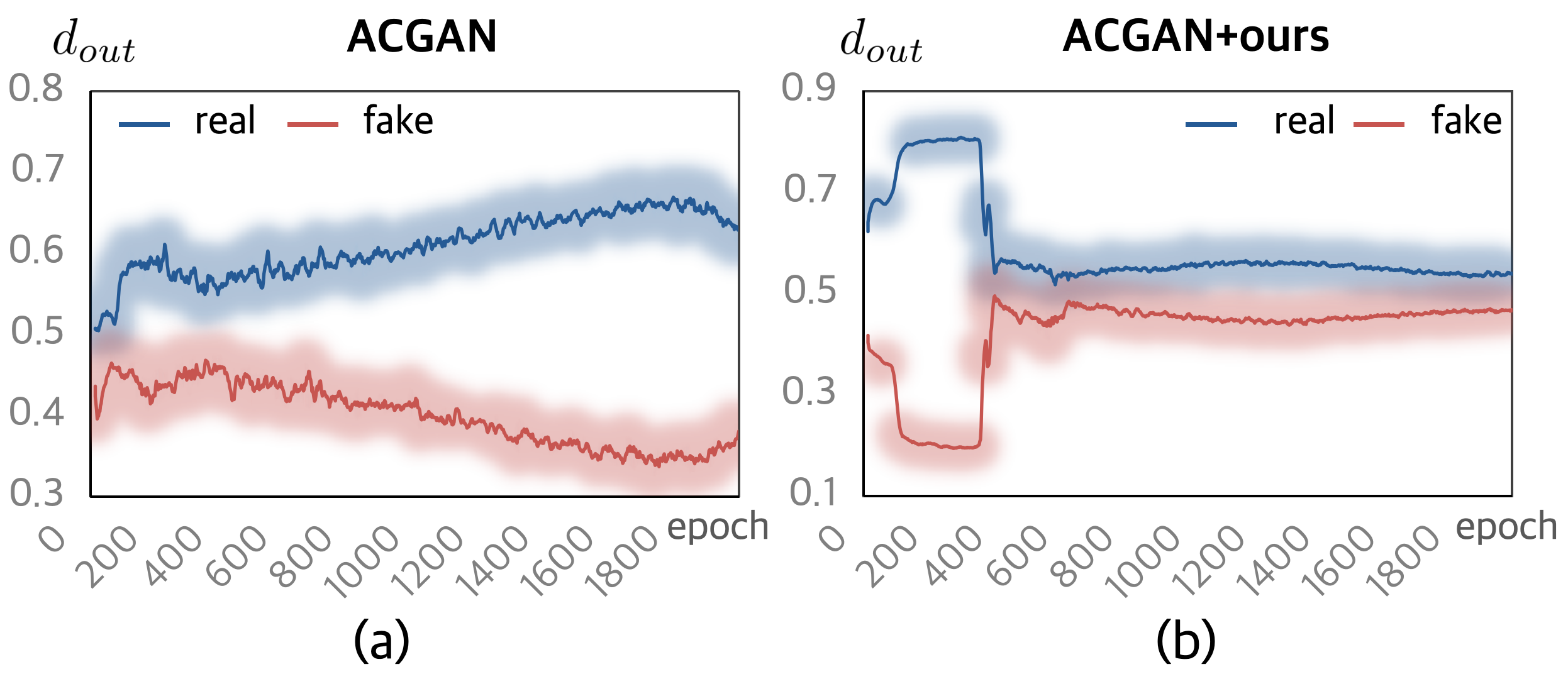}
  \caption{The discriminator output of ACGAN with and without our proposed physics-inspired regularization.}
  \label{fig:d_curve}

\end{minipage}%
\end{figure*}

\noindent \textbf{Comparisons based on ACGAN.} Tables \ref{tab:acganmstar} and \ref{tab:acganships} present a comparison of different stabilizing approaches applied to the ACGAN model on the 5\% and 10\% MSTAR datasets, as well as the 1\% OpenSARShip dataset and 14\% SAR airplane dataset. It is noteworthy that the differentiable augmentation (DA) method significantly degrades performance in SAR image generation, and other approaches also show limited effectiveness for this task. This finding confirms that natural image-oriented techniques are not well-suited for SAR data. In contrast, our method achieves superior performance, setting new state-of-the-art results across these SAR datasets.

\begin{table*}[t]
\noindent
\begin{minipage}[t!]{0.48\linewidth}
    \vspace{-0.5cm}
    \centering
    \caption{Ablation of each regularization loss.}
    \label{tab:ablaregul}
    \resizebox{\linewidth}{!}{
    \begin{tabular}{cccccc}
    \toprule
    \multicolumn{1}{c}{\textbf{Method}} & \textbf{SSIM($\uparrow$)} & \textbf{FSIM($\uparrow$)}                       & \textbf{GMSD($\downarrow$)}                       & \textbf{FID($\downarrow$)}                        & \textbf{KID($\downarrow$)}             \\
    \midrule
    \textbf{ACGAN}           & 0.3224                                 & 0.7432                                 & 0.1510                                 & 290.0484                                 & 0.4548                                 \\
    \quad \textit{w/} $\mathcal{L}_{\mathrm{phy/s}}^G$   &  0.3514 & 0.7611 & 0.1392 &  97.4472 & 0.0671 \\
    \quad   \textit{w/} $\mathcal{L}_{\mathrm{phy/f}}^G$&0.3477 &0.7525 &0.1414 & 99.9256& 0.0568\\
    \quad  \textit{w/} $\mathcal{L}_{\mathrm{phy}}^D$  & 0.3467&0.7526 &0.1422 &92.7562&0.0471\\

    \quad   \textit{w/} $\mathcal{L}_{\mathrm{phy/f}}^G$,$\mathcal{L}_{\mathrm{phy}}^D$          & 0.3489                              & 0.7515                                 & 0.1418                                 & 95.9290                                 & 0.0493                                 \\
    \quad \textit{w/} $\mathcal{L}_{\mathrm{phy/s}}^G$,$\mathcal{L}_{\mathrm{phy/f}}^G$            & 0.3357                       & 0.7586                                 & 0.1419                                 & 177.1856                                 & 0.2238                                 \\
    \midrule
    \rowcolor{gray!20}
    \textbf{Ours}       &  {\color[HTML]{D83931}\textbf{0.3583}} &  {\color[HTML]{D83931}\textbf{0.7622}} & {\color[HTML]{D83931}\textbf{0.1385}} & {\color[HTML]{D83931}\textbf{87.2719}} & {\color[HTML]{D83931} \textbf{0.0414}}\\
    \bottomrule
    \end{tabular}}
\end{minipage}%
\hfill
\noindent
\begin{minipage}[t!]{0.45\linewidth}
    \small
    \centering
    \caption{Compare the proposed physics-inspired neural module with Harris-Laplacian corner (HLC) detector.}
    \label{tab:ablaHL}
    \resizebox{0.9\linewidth}{!}{
    \begin{tabular}{ccc|cc}
    \toprule
                                      & \multicolumn{2}{c|}{\textbf{10\% MSTAR}}                                                                                            & \multicolumn{2}{c}{\textbf{5\% MSTAR}}                                                                                                                                                  \\
    \cmidrule{2-5}
    \multirow{-2}{*}{\textbf{Method}} &  \textbf{FID($\downarrow$)}                        & \textbf{KID($\downarrow$)}                     &  \textbf{FID($\downarrow$)}                        & \textbf{KID($\downarrow$)}                    \\
    \midrule
    \textbf{ACGAN}             & 290.0484                                & 0.4548                               & 340.2440                                 & 0.4713                                 \\
     \textbf{Ours(HLC)}              & 263.2789                                & 0.4146                                 & 428.3364                                 & 0.6817                                 \\
      \textbf{Ours(PSC)}                &
     
    
    {\color[HTML]{D83931} \textbf{87.2719}} &
    {\color[HTML]{D83931} \textbf{0.0414}} &
    
    
    {\color[HTML]{D83931} \textbf{130.2311}} & {\color[HTML]{D83931} \textbf{0.0861}}\\
    \bottomrule
    \end{tabular}}
\end{minipage}%
\end{table*}

\noindent \textbf{Comparisons based on StyleGAN.} 
Table \ref{tab:stylegan} compares our method with the original StyleGAN model. The results indicate that the StyleGAN baseline performs poorly in SAR image generation due to severe data scarcity. However, our approach significantly improves the model’s generative capabilities.


\noindent \textbf{Comparisons with SAR image generation methods.} 
We also compare our method with other SAR image generation approaches, including CAE \cite{guoCausalAdversarialAutoencoder2023} and CVAE-GAN \cite{huFeatureLearningSAR2021}, with the results shown in Table \ref{tab:sargen}. It is important to note that CVAE-GAN \cite{huFeatureLearningSAR2021} was not designed for data-scarce scenarios, making it challenging to train on the 10\% MSTAR dataset. However, integrating our proposed $\Phi$-GAN significantly improves its performance, demonstrating the effectiveness of our approach.

\subsection{Ablation Studies}

\noindent \textbf{Ablation of each regularization loss.} Table \ref{tab:ablaregul} discusses the effectiveness of each regularization loss design. The results show that penalizing the inconsistency of physical parameters using $\mathcal{L}_{\mathrm{phy/s}}^G$ yields the most significant performance improvement. Notably, applying $\mathcal{L}_{\mathrm{phy}}^G$ without constraining $\mathcal{L}_{\mathrm{phy}}^D$ results in a performance drop, indicating that discriminator overfitting adversely affects the alignment of generated SAR images with real electromagnetic reconstructions. Combining both feature-level constraints for the generator and discriminator, i.e., using $\mathcal{L}_{\mathrm{phy/f}}^G$ and $\mathcal{L}_{\mathrm{phy}}^D$ together, results in substantial performance gains. Overall, regularizing both the generator and discriminator at the image and feature levels using the physical model results in the greatest performance improvement.

\noindent \textbf{Ablation of physics-inspired neural module.} To highlight the effectiveness of our proposed physics-inspired neural module, we compare the results by replacing it with an off-the-shelf Harris-Laplacian corner detection method \cite{mikolajczyk2004scale}, which is widely applied in SAR image analysis to extract the scattering point. The quantitative results are demonstrated in Table \ref{tab:ablaHL}. Detailed analyses of Harris-Laplacian corner detection and our proposed method are provided in the supplementary materials.

\noindent \textbf{More analyses.} The supplementary materials include hyperparameter ablation studies and additional visualizations of SAR images generated by different approaches.

\begin{figure*}[t]
  \centering
  \includegraphics[width=0.75\linewidth]{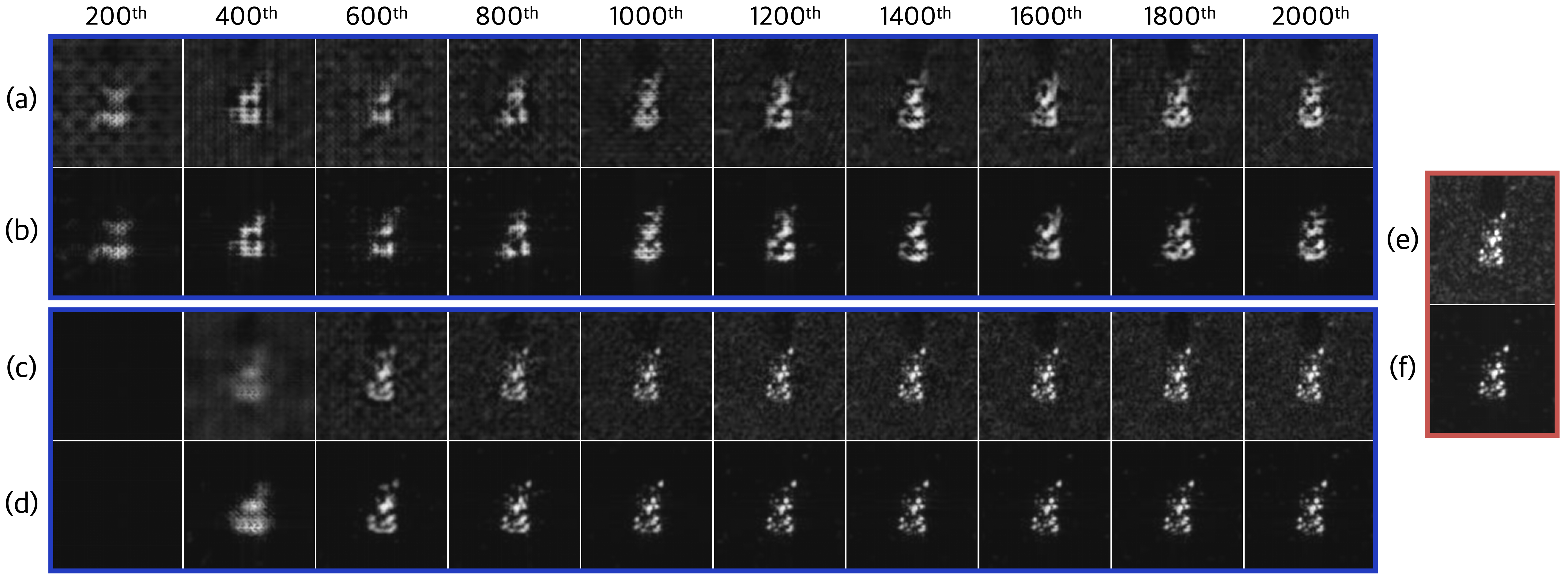}
  \caption{The generated SAR images and their associated PSC model reconstruction results during model training (from 200$^{th}$ epoch to 2000$^{th}$ epoch). (a) and (c) are the generated SAR images of ACGAN and $\Phi$-ACGAN, respectively. (b) and (d) are the PSC reconstruction results of (a) and (c), respectively. (e) and (f) are the training image and its PSC reconstruction result.}
  \label{fig:train_process}
\end{figure*}

\begin{figure*}[t]
  \centering
  \includegraphics[width=0.85\linewidth]{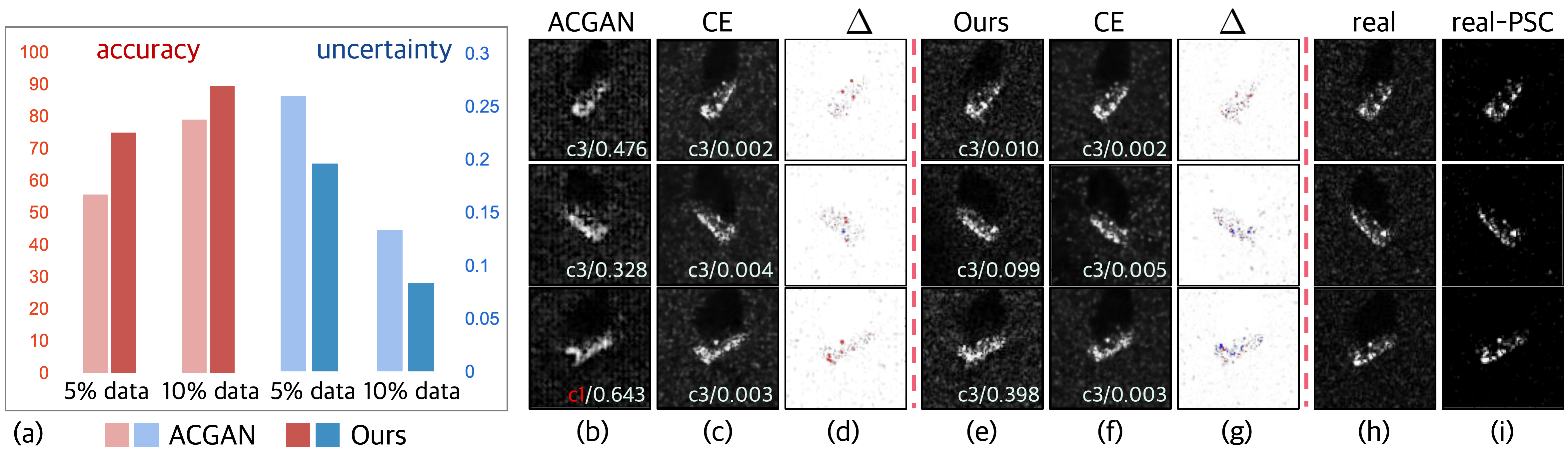}
  \caption{We evaluate the generated SAR images by using the "X-Fake" method \cite{huang2024x}. (a) Quantitative metrics reported by the Probabilistic Evaluator (Prob-Eva) of "X-Fake". (b) and (e) are generated images by ACGAN and ACGAN+ours, respectively. (c) and (f) are the counterfactual explanation results of (b) and (e), respectively. "c3/0.476" illustrates the predicted label and uncertainty by Prob-Eva. (d) and (g) are the differences between the generated SAR images and their counterfactual explanations, where the red contents illustrate the missing details of the generated images. (h) and (i) are the real SAR images with the given class and azimuth angle conditions, and their PSC reconstruction results, respectively. }
  \label{fig:counterfactual}
  \vspace{-6mm}
\end{figure*}


\subsection{Discussions}

\noindent \textbf{Stabilizing GAN training.} Figure \ref{fig:d_curve}a shows that ACGAN training with only 5\% data of MSTAR fails to converge, where the discriminator is likely overfitting to the training set and dominating the training process. As a comparison, our method significantly alleviates it and leads to gradual equilibrium, shown in Figure \ref{fig:d_curve}b. 

We demonstrate that with our proposed physical consistency regularization, the generator learns to produce outputs that align more closely with real data from a physical perspective more efficiently. As shown in Figure \ref{fig:train_process}a, the generator of baseline model fails to learn high-quality and stable scattering features given the same conditions. The PSC estimation results given in Figure \ref{fig:train_process}b show that the electromagnetic scattering features are changing during training, demonstrating a large difference with the real physical parameters shown in Figure \ref{fig:train_process}f. In comparison, our method leverages the constraint of physical parameters to guide the generator to learn the crucial scattering centers efficiently and correctly. The regularization between the latent features helps to refine the scattering details of SAR target while maintaining the physical consistency, as demonstrated in Figure \ref{fig:train_process}c and d.

\noindent \textbf{Utility evaluation.} In addition to visual quality assessment using IQA metrics, we evaluate the utility of generated SAR images in real-world applications with the “X-Fake” framework \cite{huang2024x}. This framework includes a probabilistic evaluator (Prob-Eva) and a counterfactual explainer (CE) to assess target recognition utility. Prob-Eva, a Bayesian neural network trained on real SAR images, reports higher accuracy and lower uncertainty for more realistic generated images. Figure \ref{fig:counterfactual} shows quantitative results, where our method outperforms the baseline by increasing Prob-Eva accuracy by over 20\% and significantly reducing uncertainty. Additional comparisons with other methods are provided in the supplementary materials.


\noindent \textbf{Physical consistency analysis.} Since SAR images are difficult for visual interpretation, we leverage the counterfactual explainer (CE) of "X-Fake" to help us investigate the electromagnetic scattering details of the generated data. CE can generate the counterfactual image with inauthentic details of the input SAR image. As demonstrated in Figure \ref{fig:counterfactual}, the differences between generated SAR image and CE results are computed as $\Delta$, where the absent scattering details are highlighted in red. We can compare $\Delta$ with real SAR images and their PSC reconstruction results, showing that the generated images of the baseline model lack more details highlighted by the physical parameters of the real SAR images. The results further prove the advantage of maintaining the physical consistency of our method.

\section{Conclusions}

For SAR image generation, we propose a physics-inspired regularization method, termed $\Phi$-GAN, to enhance GAN training. By integrating the electromagnetic physical model directly into the GAN framework in an end-to-end manner, $\Phi$-GAN ensures that the generated SAR targets align with established physical principles and prevents the discriminator from overfitting. $\Phi$-GAN outperforms existing data augmentation and regularization techniques in stabilizing GAN training and improving generalization with limited training samples. It achieves superior results compared to state-of-the-art methods across various GAN models and SAR image datasets.

\noindent \textbf{Limitation and future work.} While advanced methods such as diffusion models and NeRF have shown impressive results with natural images, their application to SAR images is still limited. This work represents an initial effort to incorporate physical models into GANs for SAR image generation. Future research will focus on extending these physical models to more advanced generative techniques.
{
    \small
    \bibliographystyle{ieeenat_fullname}
    \bibliography{main}
}


\end{document}